\documentclass{article}

\usepackage[preprint]{neurips_2021}
\setcitestyle{square}

\usepackage[utf8]{inputenc} 
\usepackage[T1]{fontenc}    
\usepackage{hyperref}       
\usepackage{url}            
\usepackage{booktabs}       
\usepackage{amsfonts}       
\usepackage{nicefrac}       
\usepackage{microtype}      
\usepackage{xcolor}         
\usepackage[export]{adjustbox}
\usepackage{caption, subcaption, floatrow}
\title{Fix your Models by Fixing your Datasets}

\author{Atindriyo Sanyal \\Galileo Technologies \\ \texttt{atin@rungalileo.io} \And Vikram Chatterji \\Galileo Technologies\\ \texttt{vikram@rungalileo.io} \And Nidhi Vyas \\Galileo Technologies\\ \texttt{nidhi@rungalileo.io} \And Ben Epstein \\Galileo Technologies\\ \texttt{ben@rungalileo.io} \And Nikita Demir \\Galileo Technologies\\ \texttt{nikita@rungalileo.io} \And Anthony Corletti \\Galileo Technologies\\ \texttt{anthony@rungalileo.io}  }

\begin{document}

\maketitle

\begin{abstract}
The quality of underlying training data is very crucial for building performant machine learning models with wider generalizabilty. However, current machine learning (ML) tools lack streamlined processes for improving the data quality. So, getting data quality insights and iteratively pruning the errors to obtain a dataset which is most representative of downstream use cases is still an ad-hoc manual process. Our work addresses this data tooling gap, required to build improved ML workflows purely through data-centric techniques. More specifically, we introduce a systematic framework for (1) finding noisy or mislabelled samples in the dataset and, (2) identifying the most informative samples, which when included in training would provide maximal model performance lift. We demonstrate the efficacy of our framework on public as well as private enterprise datasets of two Fortune 500 companies, and are confident this work will form the basis for ML teams to perform more intelligent data discovery and pruning. 



\end{abstract}

\section{Introduction}
Machine learning is key to mission critical data-driven decisions across an array of applications. The primary goal of model training phase is improving overall metrics like F1 and RMSE. A common data-centric technique to achieve this goal is adding more samples and features to the dataset. However, there are multiple issues with the over-use of this approach. First, the benefits of strictly increasing dataset sizes is uncertain. Instead, adding lesser but more informative samples to training or cleaning noisy samples from a dataset has provided significant model improvement \cite{axelrod-etal-2015-class, dou2020dynamic, xu-koehn-2017-zipporah}. Second, training complexity increases exponentially when more features are added to the model viz. `curse of dimensionality'. This results in limited generalizability when using a fixed budget to add new samples.

A direct repercussion of such biased models in production is that of `model downtimes', where the model makes frequent incorrect predictions. Debugging and alleviating these downtimes require large hours of manual effort. As the ML footprint of an organization grows, this process becomes increasingly untenable. This is where data-centric ML tooling for automatically detecting and surfacing these errors grows considerably.

Many machine learning teams across industries agree that \textit{data quality} is the key to produce more robust and generalizable models. Yet, there is a lack of streamlined tools and processes to automate data cleaning workflows (i.e. to prune out noisy samples), and optimizing data labelling for fixed budget scenarios (i.e. to selectively add most informative samples). Instead, the common approach is largerly ad-hoc, leading to enormous waste of human capital during the process of uncovering causes for poor model performance.

The goal of our work is to provide a framework which uses systematic data-centric methods to improve data quality across the ML workflow, thereby improving model performance. The first approach automatically identifies labelling errors in a given dataset. We provide a data-dependent and a data-independent approach to surface these noisy labels. The second approach identifies most informative samples in an unlabelled set, which can be used for training a model to achieve maximal model performance lift. This approach can improve model robustness with minimal labelling effort/budget. Our work can produce higher quality models, smaller but better quality datasets (saving model training time), quicker but informed dataset labelling (saving annotation cost).

To demonstrate the overall adaptability of our framework, we provide results for text classification on two public and two private enterprise datasets. Our methods achieve significant improvements over baselines, and are a proof-of-concept for practical dataset engineering using data-centric techniques for producing and maintaining high quality models in production.

\section{Improving ML Dataset Quality}

Data labelling is a laborious and expensive process. Enterprises that have terabytes of unlabelled data often need to optimize for labelling a subset of data that will achieve the largest performance lift. Since it involves significant human intervention, these labels can still be susceptible to many errors\footnote{https://labelerrors.com/}. 
In this section, we describe two data-centric methodologies employed to identify errors in ground truth data, and a subset of informative samples for maximal performance lift.

\subsection{Identifying Labelling Errors}

\label{sec:label_errors}
\textbf{Model based confidence-certainty metric} is a data-dependant approach derived from prior work on \textit{Dataset Cartography} \cite{swayamdipta2020dataset}, and automatically identifies noisy/mislabelled samples from a dataset. Our variation to prior work combines prediction probability of ground truth and overall prediction distribution across labels to score every sample in the dataset. More precisely, we use \textit{model confidence} (\(\mu\)), where \(\mu_i\) is the prediction probability of ground truth, and \textit{model certainty} (\(\delta\)), where \(\delta_i\) is the difference between the prediction probability of argmax and next argmax (i.e. margin distribution) for a sample \(x_i\). \(\mu\) captures how confident the model is while predicting ground truth and \(\delta\) captures how certain the model is while making a prediction. Unlike \cite{swayamdipta2020dataset}, we only use scores from the penultimate epoch of training. For each training sample \(x_i\), we first compute \(\mu_i\) and \(\delta_i\) and then partition the data into segments of high/low confidence-certainty scores. The 90th percentile of datums from the least confident and most certain score range are marked as noisy/mislabelled.

\textbf{Model based certainty metric} is a data-independent approach derived from prior work on \textit{Confident Learning} (CI) \cite{northcutt2021confident}\footnote{https://github.com/cleanlab/cleanlab}, and automatically learns a class-conditional joint distribution (Q) between provided dataset labels (assumed noisy) and the latent labels (assumed uncorrupted), to identify noisy samples. This approach presumes each class may be independently mislabelled as an incorrect class with a probability that is data-independent. At inference, Q is used to compute a \textit{model certainty} (\(\delta\)) score for every sample, and the top 90 percentile with highest scores are marked as noisy/mislabelled.

Both approaches described above are highly generalizable across other domains as they require minimal code-injection to the underlying model and the associated loss function is kept unchanged. However, scaling and automating them in real-world applications is still a challenge (refer \textsection
 \ref{sec:challenge}).

\subsection{Identifying Most Informative Samples}
In production workflows, it is computationally impractical to use traditional active learning strategies that select a single sample at each step, and retrain the model each time. This approach uses prior work of
\textit{Core-set} \cite{sener2018active}, where given a fixed annotation budget, the goal is to select a batch subset of \textit{B} samples that would provide maximal model performance lift. More precisely, the geometry of datapoints is used to pick \textit{K} well-distributed cores, and these cores are then used to pick \textit{B} samples. The approach optimizes picking a subset that will be competitive for the remaining datapoints. So, given an initial set of samples and their embeddings, {core-set} selection returns a subset of samples that should be used to expand the training data.

\section{Experiments and Results}
\label{sec:experiments}
We validated our framework on text classification using public (Newsgroup20\footnote{http://qwone.com/~jason/20Newsgroups/} and Toxicity\footnote{https://www.kaggle.com/c/jigsaw-toxic-comment-classification-challenge/overview}) and private enterprise (two Fortune 500 FinTech company) datasets. All datasets are multi-class, except Toxicity dataset which is two-class. We add a top-layer to a pretrained DistilBERT model from Hugging Face for all our experiments\footnote{https://huggingface.co/transformers/model\_doc/distilbert.html}and fine-tune using early-stopping (min delta 0.001 on training accuracy).

\textbf{A. Identifying Labelling Errors}: We manually induce 10\% noise by randomly flipping labels in the two public datasets (i.e. setup similar to \cite{swayamdipta2020dataset}). For private enterprise datasets (which correspond to real-world production data), we re-labelled the samples via MTurk\footnote{https://www.mturk.com/}, and found labelling errors in 1.5\% and 33\% of the input data, respectively. We observed that samples in the lower confidence and higher certainty region are further away from decision boundary, and correspond to hard or mislabelled training samples. The model found it difficult to learn from these samples potentially because they are noisy (refer Table 1 for examples).

As shown in Table 2 (left), both methods are very efficient at finding mislabelled samples. Since \textit{certainty-metric} adopts a data-independent approach, it outperforms \textit{confidence-certainty-metric} on public datasets where the labels were randomly flipped (and are independent of the underlying samples). Another expected outcome was lower gains on private datasets, mainly because a systematic human mislabelling error is harder to identify than artificial random noise. Since such errors are more data-dependant, we hypothesize that \textit{confidence-certainty-metric} will outperform other method.

\textbf{B. Identifying Informative Samples}: We used the Toxicity multilabel dataset and trained a baseline model on 3 sets of 1000 initial seed samples: randomly sampled, selected from the decision boundary, and selected furthest from the decision boundary. Next, we employed three different strategies to label and expand this training data (1) Random sampling where we pick new samples randomly without replacement (2) Certainty based sampling where we use the baseline model to pick the batch of samples with highest certainty scores (3) Core-set based sampling where we use the baseline model to extract embeddings and use core-set recommendation to pick top candidates. Note, for this experiment, we set the annotation budget to 300 samples, so the final bench-marked model is trained on 1300 training samples. 

As shown in Table 2 (right), both core-sets and certainty based methods perform better than random sampling. Core-sets outperform certainty when the initial seed samples were selected away from decision boundary (or at random) and did not include important support vectors that determine the overall decision boundary. By the virtue of this method, we are able to select more diverse samples that capture the support (see Figure 1 (right)). More generally, certainty based approaches select multiple samples from the same region (see Figure 1 (middle)) and so the model misses out on insights that a core-set would discover from its broader coverage. Given equal performance lifts between these two methods, core-set provides the benefit of a more diverse selection and thus perhaps a more robust model.

\begin{table}
\small
\begin{tabular}{@{}lll@{}}
\toprule
\textbf{Sample}                                                                                                                                                                                                           & \textbf{Noisy Label}  & \textbf{Correct Label}   \\ \midrule
\begin{tabular}[c]{@{}l@{}}It is a long standing good luck \\ Redwing's tradition to throw an octopus \\ on the ice during a Stanley Cup game\end{tabular}                                                               & comp.graphics      & rec.sports.hockey \\\midrule
\begin{tabular}[c]{@{}l@{}}Excuse me? This has *already* happened. \\ There's a couple of humps in the tent already. \\ Ask the folks at Qualcomm what became of the \\ non-trivial encryption scheme\end{tabular}          & talk.politics.misc & sci.electronics   \\ \midrule
\begin{tabular}[c]{@{}l@{}}All these people who send in their polls should \\ take a closer look at NJD, they are a very \\ deep team, with two very capable goalies, \\and excellent forwards and defensemen\end{tabular} & comp.graphics      & rec.sports.hockey \\\bottomrule
\end{tabular}
\caption{Top mislabelled samples picked by confidence-certainty method}
\end{table}

\begin{table}[!htb]
    \caption{Results: \textbf{(Left)} identifying dataset labelling errors \textbf{(Right)} identifying most informative samples (NW=Newsgroup20, TOX=Toxicity, Anon1 and Anon2 are private enterprise datasets; M1=confidence-certainty method, M2=certainty method, Rand=Random, DB=Decison Boundary)}
    \begin{minipage}{.6\linewidth}
      \centering
      \small
        \begin{tabular}{@{}lllll@{}}
         \toprule
                                     & \textbf{NW} & \textbf{TOX} & \textbf{Anon1} & \textbf{Anon2} \\ \midrule
\#Mislabeled (induced)                &    2054        &          2051        &          -       &         -        \\
\#Mislabeled (MTurk) & - & - & 412 & 117 \\
\#Mislabeled (M1)                 &       2032      &         2041          &          382       &          101       \\
\#Mislabeled (M2) & 2039 & 2045  & - & -\\ \midrule
\%Accuracy (M1)                 &     98.9 & 99.5 & 92.7 & 86.3 \\
\%Accuracy (M2)                 &     \textbf{99.3} & \textbf{99.7} & - & - \\

 \bottomrule
        \end{tabular}
    \end{minipage}%
    \begin{minipage}{.4\linewidth}
      \centering
      \small
        \begin{tabular}{@{}llll@{}}
\toprule
 & Rand. & DB & Not DB \\ \midrule
\textbf{Baseline} & 0.42 & 0.34 & 0.08 \\ \midrule
\textbf{Random} & 0.44 & 0.42 & 0.42 \\
\textbf{Certainty} & \textbf{0.52} & \textbf{0.52} & 0.49 \\
\textbf{Core-set} & \textbf{0.52} & 0.50 & \textbf{0.51}\\ \bottomrule
 
\end{tabular}
    \end{minipage} 
\end{table}

\begin{figure}
\centering
\minipage{0.32\textwidth}
  \includegraphics[width=\linewidth]{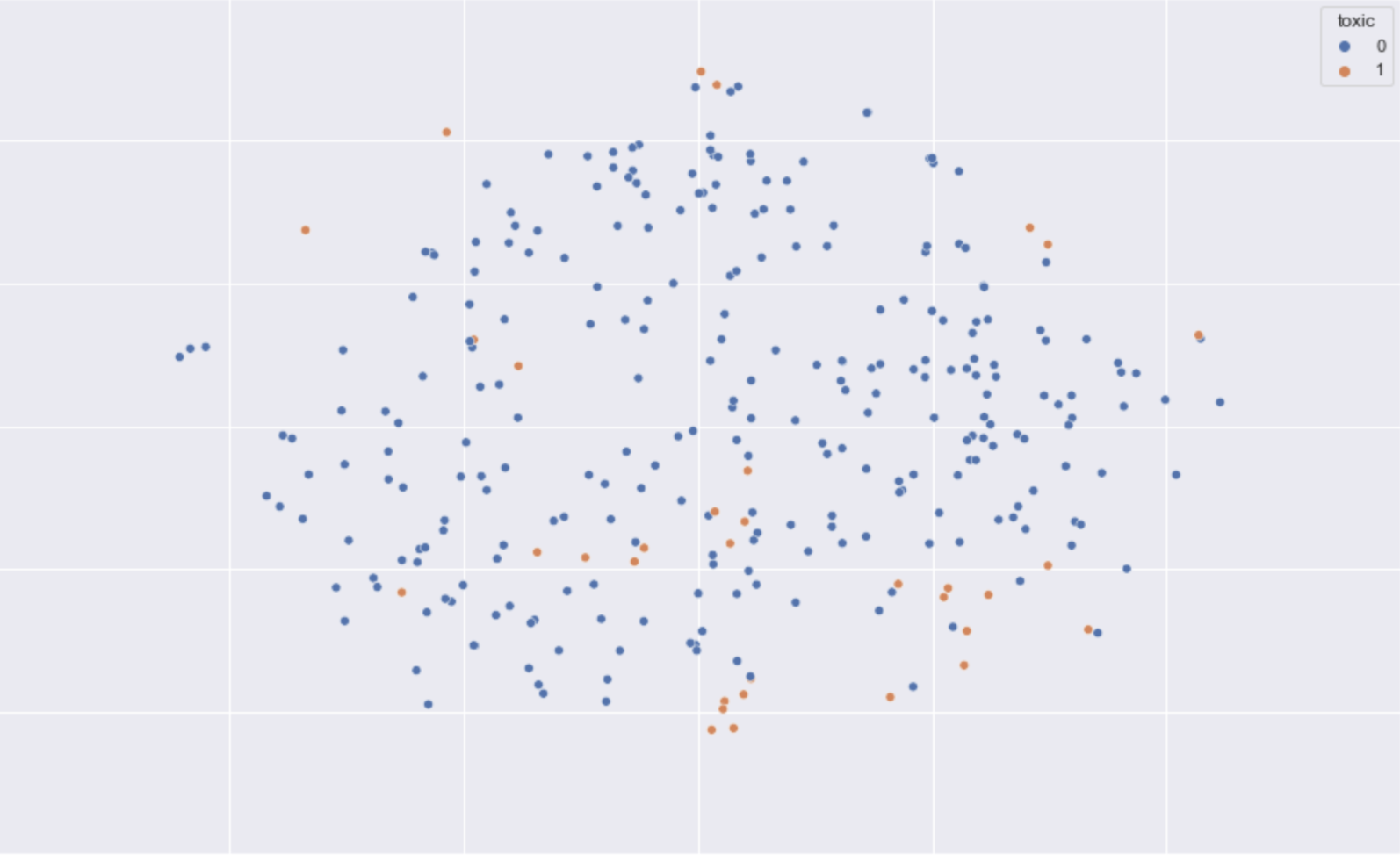}
\endminipage\hfill
\minipage{0.32\textwidth}
  \includegraphics[width=\linewidth]{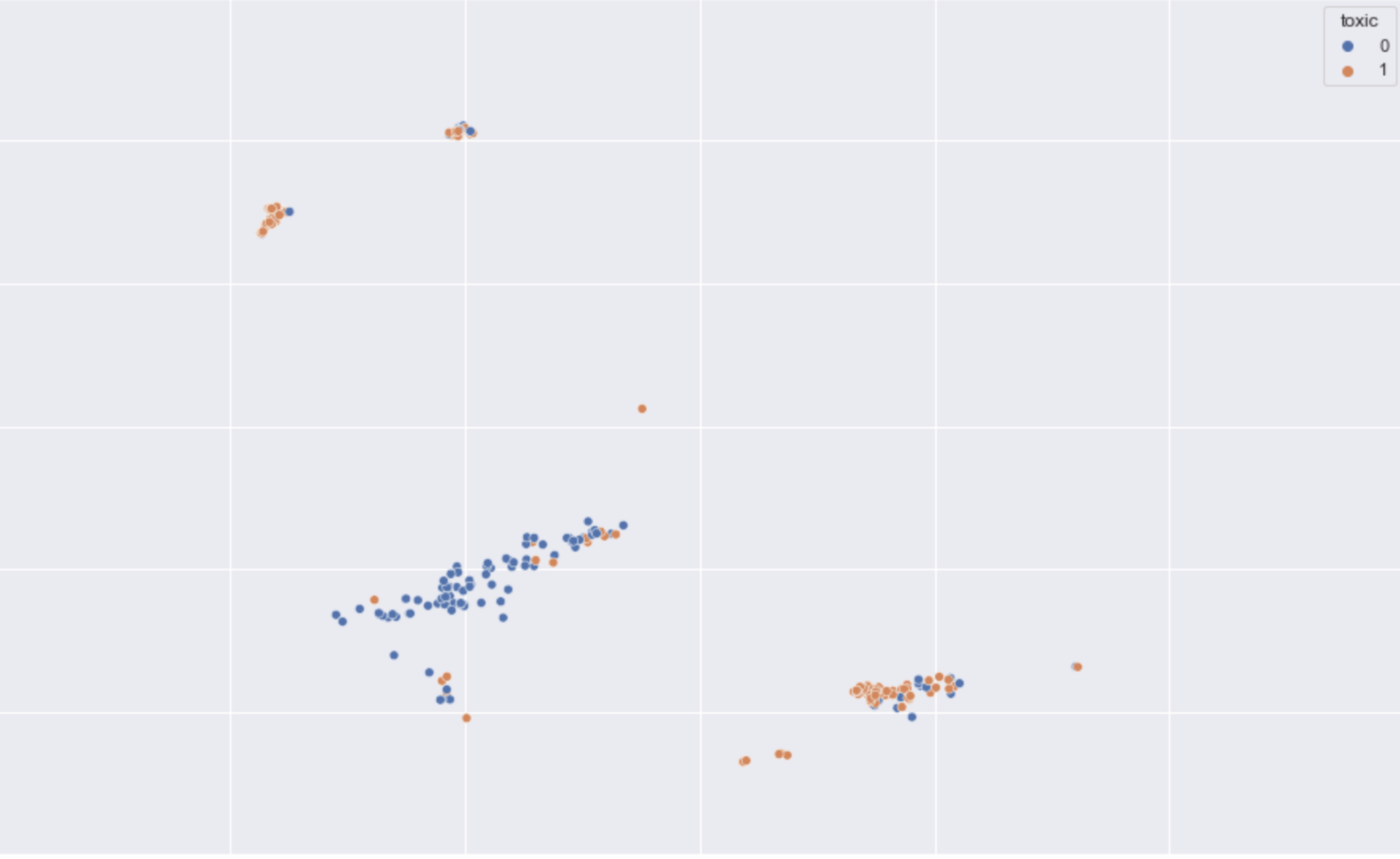}
\endminipage\hfill
\minipage{0.32\textwidth}%
  \includegraphics[width=\linewidth]{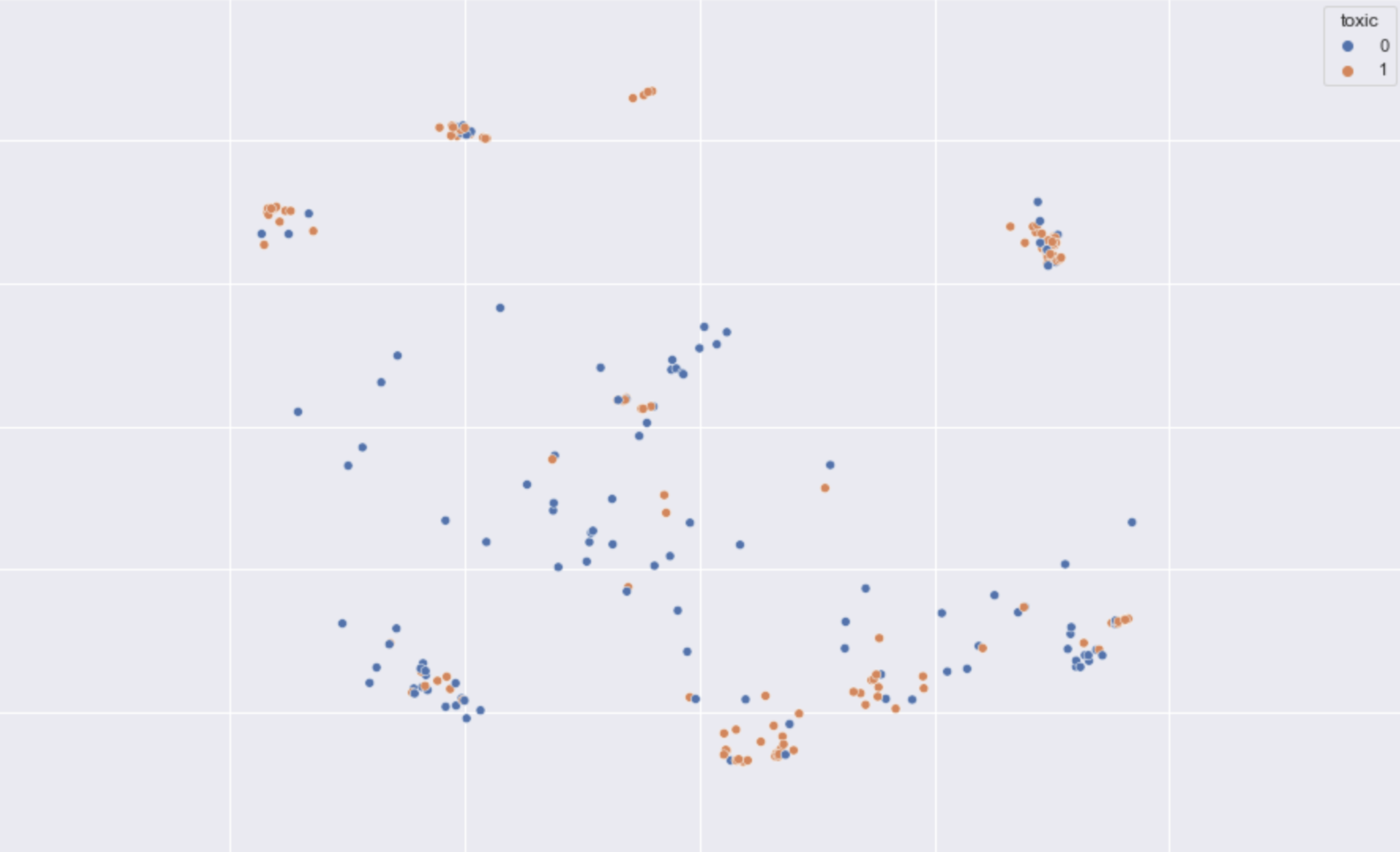}
\endminipage
\caption{Samples picked by random sampling (Left) certainty sampling (Middle) core-set sampling (Right) across unlabelled data. Note the unseen cluster core-set captures in the upper-right corner.}
\label{fig}
\end{figure}

\section{ML Data Observability in Production}
\label{sec:challenge}
Identifying informative samples and labelling errors are just two out of many other problems and challenges which fall under the umbrella of \textit{ML Data Observability}. In practice, implementing these techniques have many challenges, and we discuss one potential solution in this section. 

Over the last few years, parts of the ML infrastructure pipeline have been standardized such as \textit{feature stores} \cite{orr2021managing} and \textit{model SDKs}, which enable engineers to build and maintain ML models with ease. As a result, there has been an explosion of models being trained and deployed for various business applications. Yet, errors in real-time production data is extremely common due to lack of robust data quality solutions. For instance, there is no standard set of libraries that provide theoretical guarantees, and most workflow systems provide tooling for monitoring model level metrics, which is not sufficient for guaranteeing robust models. In addition, present-day approaches to data observability is highly manual and requires significant engineering.

An API-first approach could be one way to provide this required tooling for implementing such data centric approaches quickly. Such an approach would provide easy integration with existing bespoke ML infrastructure workflows, and allow ML practitioners to train more robust models quickly through key data-centric insights allowing companies to iterate quickly on improving models and reduce potential bias in data points. 


For ML organizations to succeed, it's critical to bake data quality into their ML ecosystems. As models become more pervasive, it becomes harder to know when things go wrong; be it at the model, data or systems level. So, automating data quality using data-centric techniques is key to root-causing and solving these issues, ensuring models run at high quality consistently in production.

\begin{ack}
We acknowledge the support of \textbf{Stanford AI Lab} (SAIL) for their useful feedback and discussions that helped improve this work.
\end{ack}

\newpage
 {\small
 \bibliographystyle{plainnat}
 \bibliography{neurips_2021}
 }


\end{document}